\documentclass{article}

\PassOptionsToPackage{numbers, sort&compress}{natbib}
\usepackage[preprint]{neurips_2021}

\usepackage[utf8]{inputenc} %
\usepackage[T1]{fontenc}    %
\usepackage{hyperref}       %
\usepackage{url}            %
\usepackage{booktabs}       %
\usepackage{amsfonts}       %
\usepackage{nicefrac}       %
\usepackage{microtype}      %
\usepackage{xcolor}         %
\usepackage{verbatim}

\usepackage{graphicx}
\usepackage{wrapfig}
\usepackage{subcaption}
\usepackage{tabularx}
\usepackage{sidecap}

\sidecaptionvpos{figure}{c}

\title{C5T5: Controllable Generation of Organic Molecules with Transformers}

\newif\ifcomments
\commentsfalse
\ifcomments
\newcommand{\joey}[1]{\textbf{\textcolor{magenta}{joey: #1}}}
\newcommand{\ujval}[1]{\textbf{\textcolor{blue}{ujval: #1}}}
\newcommand{\art}[1]{\textbf{\textcolor{teal}{alex: #1}}}
\else
\newcommand{\ujval}[1]{\textcolor{black}{}}
\newcommand{\joey}[1]{}
\newcommand{\art}[1]{}
\fi

\definecolor{greenish}{RGB}{64,176,166}
\newcommand{\ch}{{\color{blue}\checkmark}}
\newcommand{\x}{{\color{red}\textbf{$\mathcal{X}$}}}

\author{%
  Daniel Rothchild \\
  EECS \\
  UC Berkeley\\
  \texttt{drothchild@berkeley.edu} \\
  \And
  Alex Tamkin \\
  Computer Science \\
  Stanford University \\
  \texttt{atamkin@stanford.edu} \\
  \And
  Julie Yu \\
  Data Science \& Chemistry\\
  UC Berkeley \\
  \texttt{julieyu@berkeley.edu} \\
  \And 
  Ujval Misra \\
  EECS \\
  UC Berkeley \\
  \texttt{ujval@berkeley.edu} \\
  \And
  Joseph Gonzalez \\
  EECS \\
  UC Berkeley \\
  \texttt{jegonzal@berkeley.edu}
}

\begin{document}

\maketitle

\begin{abstract}

Methods for designing organic materials with desired properties have high potential impact across fields such as medicine, renewable energy, petrochemical engineering, and agriculture.
However, using generative modeling to design substances with desired properties is difficult because candidate compounds must satisfy multiple constraints, including synthetic accessibility and other metrics that are intuitive to domain experts but challenging to quantify.
We propose C5T5, a novel self-supervised pretraining method that enables transformers to make zero-shot select-and-replace edits, altering organic substances towards desired property values.
C5T5 operates on IUPAC names---a standardized molecular representation that intuitively encodes rich structural information for organic chemists but that has been largely ignored by the ML community.
Our technique requires no edited molecule pairs to train and only a rough estimate of molecular properties, and it has the potential to model long-range dependencies and symmetric molecular structures more easily than graph-based methods.
C5T5 also provides a powerful interface to domain experts:
it grants users fine-grained control over the generative process by selecting and replacing IUPAC name fragments, which enables experts to leverage their intuitions about structure-activity relationships.
We demonstrate C5T5's effectiveness on four physical properties relevant for drug discovery, showing that it learns successful and chemically intuitive strategies for altering molecules towards desired property values.

\end{abstract}

\begin{figure}[h]
    \centering
    \includegraphics[width=0.7\linewidth]{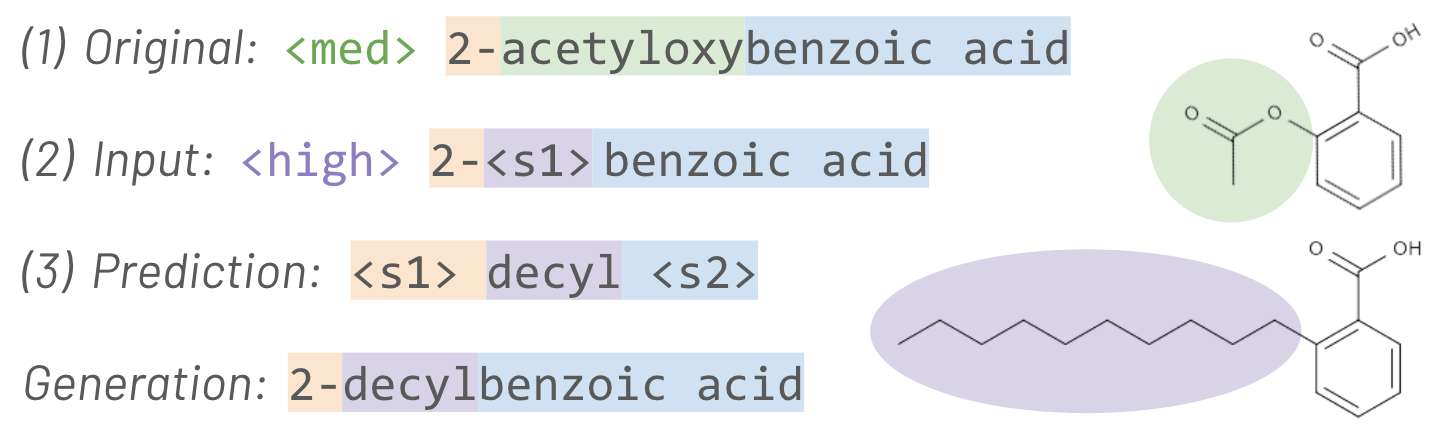}
    \caption{\textbf{Increasing a molecule's octanol-water partition coefficient with C5T5}. 
    (1) A molecular fragment (\texttt{acetyloxy}) is identified in a molecule of interest. 
    (2) The molecular fragment is replaced with a sentinel token (\texttt{<s1>}) and the property value is set to the desired bucket (\texttt{<high>}).
    (3) Sampling from C5T5 produces a new fragment (\texttt{decyl}). 
    Substituting in the fragment yields a new molecule. The long chain of carbons added to the molecule increases its solubility in octanol while decreasing its solubility in water.
    }
    \label{fig:objective}
\end{figure}

\section{Introduction}

Organic molecules are used in countless applications across human society: as medicines, industrial chemicals, fuels, pesticides, plastics, television screens, solar cells, and many others.
Traditionally, new molecules are designed for particular tasks by hand, but the space of all possible molecules is so vast (e.g. the total number of drug-like molecules may be as high as $10^{60}$) that most useful materials are probably still undiscovered \cite{enumerate-wircms12}.
To automate materials discovery, domain experts have turned to high-throughput screening, in which a large library of potentially useful molecules is generated heuristically, and the most promising molecules are chosen for further study using computational models that estimate how effective each substance will be for the target application \cite{principles-bjp11}.
Unfortunately, high-throughput screening faces the fundamental limitation that the total number of molecules that can be screened is still only a tiny fraction of all possible molecules.

Generating molecules directly using machine learning addresses this limitation, but \emph{de novo} molecular design using machine learning can be of limited use in domains like drug discovery, where experts' intuitions about structure-activity relationships and external factors like patentability are important to consider in the design process.
These constraints can often be expressed by providing known portions of the molecular structure; for example a domain expert may be interested in a particular scaffold because it has favorable intellectual property attributes, or certain parts of a drug may be needed for the desired biological activity, while other parts can be modified to increase bioavailability.

To address this real-world setting, we consider the problem of learning to make localized modifications to a molecule that change its physical properties in a desired way.
We propose C5T5: Controllable Characteristic-Conditioned Chemical Changer with T5 \cite{t5}, a novel method for generative modeling of organic molecules that gives domain experts fine-grained control over the molecular optimization process while also providing more understandable predictions than prior methods (Figure \ref{fig:objective}).
Our two key contributions are 
1) recasting molecular modeling as language modeling on the semantically rich  \textbf{IUPAC name} base representation, 
and 2) the development of a novel \textbf{conditional language modeling} strategy using transformers that supports targeted modifications to existing molecules.

\begin{SCfigure}
    \caption{\textbf{Visual representations of IUPAC names and SMILES.} Tokens in IUPAC names correspond to well-known functional groups and moieties. In contrast, tokens in SMILES correspond to individual atoms and bonds.}
    \includegraphics[width=0.5\textwidth]{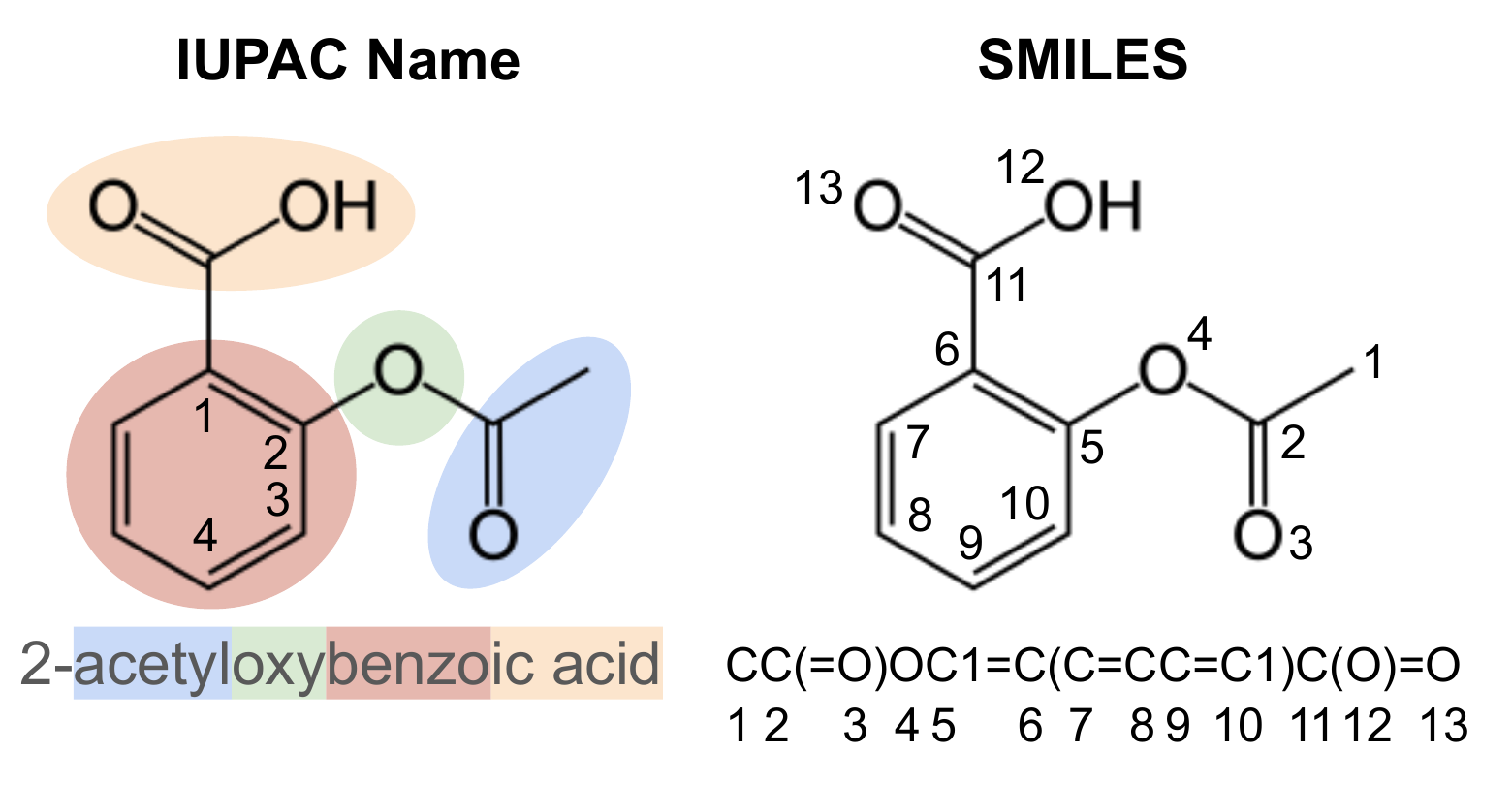}
    \label{fig:naming}
\end{SCfigure}

\paragraph{IUPAC Names.} The IUPAC naming system is a systematic way of naming organic molecules based on functional groups and moieties, which are commonly occurring clusters of connected atoms that have known chemical behaviors.
Organic chemists have discovered countless chemical reactions that operate on functional groups, and they use these reactions to develop synthesis routes for novel molecules.
Despite this, existing generative methods for organic molecules have ignored IUPAC names as a representation, instead opting for atom-based representations like SMILES \cite{smiles} and molecular graphs \cite{molfingerprints-neurips15}. We argue that these representations are less suitable for molecular optimization because adding or removing arbitrary atoms has no intuitive meaning to chemists and is unlikely to allow for easy synthesis; see Figure \ref{fig:naming} for a comparison of IUPAC names and SMILES. 
To the best of our knowledge we are the first to use IUPAC names as a base representation for molecular modeling.

\paragraph{Self-Supervised Objective for Zero-Shot Editing.} To enable targeted modifications of molecules without predefined edit pairs,  we train transformers with a conditional variant of a self-supervised infilling task by masking out certain tokens in the IUPAC name and training the model to replace these missing tokens.
Crucially, we prepend the IUPAC names with discretized property values during training, enabling the model to learn the conditional relationships between the property value and molecular structure. During inference, we replace the true property values with desired property values, mask out the portion of the molecule we want replaced, and ask the model to infill the masked tokens as usual.
To the best of our knowledge, this sort of self-supervision to enable guided select-and-replace editing has not been explored previously; we anticipate this method could be broadly applied in other controlled generation contexts, such as modeling various aspects of text like affect, politeness, or topic \citep{ghosh2017affect, Ficler2017ControllingLS, niu2018polite, Keskar2019CTRLAC}.
  
As we show in Section \ref{sec:results}, C5T5 is able to make interpretable targeted modifications to molecules that lead to desired changes across several physical properties important in drug design.

\section{Related Work}

\paragraph{Modeling.} A number of machine learning methods have been developed for the task of designing organic molecules, but most do not allow a user to make targeted modifications to a molecule.
Some methods, like generative adversarial networks and unconditional sequence models, provide no control over a generated molecule's structure 
\citep{bimodal-lstm-jcim20,organ-arxiv17,sanchez2017optimizing,molgan-icmlw18,molhyper-vae-icml19}, and are therefore more useful for generating candidate libraries than optimizing a particular molecule.
Other methods, like variational autoencoders or sequence models that are conditioned on a base molecule, allow specifying that generated molecules should be similar to a starting molecule in some learned space, but there is no way to specifically target a certain part of the molecule to modify
\citep{mmp-transformer-jcheminfo21,vjtnn-iclr19,cmg-transformer-chil21,yang2020improving,crnn-nature-mi20,bombarelli-acs-cs18,condvae-jcheminfo18,vae-transformer-cs21,cgvae-neurips18,jtvae-icml18,cyclegan-jchi20,olivecrona2017molecular,liggpt-chemrxiv21,gcpn-neurips18,graphaf-iclr20}.
Recognizing the importance of leveraging domain experts' intuition about structure-activity relationships, several methods, published mostly in chemistry venues, have explored constraining generated molecules to contain a scaffold, or a subgraph of the full molecular graph \citep{deepscaffold-jcim19,scaffold-based-vae-rsc-cs20,scaffold-based-gvae-arxiv21}.
However, these methods append to scaffolds arbitrarily instead of allowing domain experts to specify which part of the molecule they would like to modify or append to, limiting their utility for human-in-the-loop molecular optimization.

A few methods have explored allowing targeted modifications, where a domain expert can mask out a portion of a starting molecule and ask the model to replace the mask with a novel side chain
\citep{scaffold-based-lstm-attn-jcheminfo20,scaffold-constrained-rnn-jcim20,mmp-transformer-iclrw21}.
These methods are limited because they only support masking parts of the molecule that can be truncated by cutting a single bond, and because they require a dataset of paired molecules (scaffolds \& decorators) that must be constructed using hand-crafted rules.
In contrast, C5T5 learns in an entirely unsupervised fashion and therefore requires no paired data; the only limit to what can be masked is what can be represented using IUPAC tokens.

\paragraph{Representation.} Existing methods all use SMILES (or a derivative representation) or graphs to represent molecules.
There are a number of drawbacks to using the SMILES representation: a small change in a molecule can lead to a large change in the SMILES string \cite{jtvae-icml18}; flattening the graph into a list of atoms artificially creates variable- and long-range dependencies between bonded atoms; and it is difficult to reason about common substructures, because the same structure can be represented in many different ways depending on how the graph was flattened.
And although graphs seem like a natural representation for molecules, graphs do a poor job encoding symmetry, long-range interactions between atoms that are many bonds apart but nearby in 3D space, and long-range interactions that arise from conjugated systems \cite{molfingerprints-neurips15}.
C5T5 operates instead of IUPAC names, which we argue in Section \ref{sec:method_iupac} is a more suitable representation for molecular optimization because tokens have much more semantic meaning. 
See Appendix \ref{app:related} for more details on how C5T5 relates to prior work.

\paragraph{Transformers for Molecular Modeling}
Outside of molecular optimization, transformers have found a number of applications in molecular modeling tasks, 
including property prediction \cite{smiles-bert-bcb19, grover-neurips20}, 
chemical reaction prediction \cite{reaction-pred-acs-cs19},
retrosynthesis \cite{retrosynthesis-transformer-icann19} and
generating proteins \cite{prottrans-biorxiv20, protein-transformer-nature-sr21}.
A few works have explored using transformers for generative modeling of organic molecules \cite{mmp-transformer-jcheminfo21, cmg-transformer-chil21, vae-transformer-cs21}. Some works have also proposed using transformers for scaffold-conditioned generative modeling \cite{mmp-transformer-iclrw21, liggpt-chemrxiv21}.
This work extends these efforts by proposing a simple yet effective training and zero-shot adaptation method, and by using IUPAC names instead of SMILES strings.

\paragraph{IUPAC Names} Although we are unaware of prior work using IUPAC names as a base representation for molecular modeling, several works have explored using machine learning to convert between IUPAC names and other molecular representations \cite{rajan2021stout,handsel2021translating,krasnov2021struct2iupac}.

\section{Method}
\label{sec:method}
Molecular optimization is a difficult problem because it requires modifying a molecule that already satisfies a number of requirements.
Modifications need to improve a particular aspect of the molecule without degrading its performance on other metrics, and without making it too difficult to synthesize.
We argue that by using IUPAC names (Section \ref{sec:method_iupac}) and by allowing users to target particular parts of a molecule to modify (Section \ref{sec:method_t5}), C5T5 has the potential to support human-in-the-loop molecular editing that complements domain experts' intuitions about structure-activity relationships and synthetic accessibility.%

\subsection{IUPAC Naming}
\label{sec:method_iupac}
The International Union of Pure and Applied Chemistry (IUPAC) publishes a set of rules that allow systematic conversion between a chemical structure and a human-readable name \cite{favre2013nomenclature}.
For example, 2-chloropentane refers unambiguously to five carbons (``pent'') connected by single bonds (``ane'') with a chlorine atom (``chloro'') bonded to the second carbon from one end (``2-'').
IUPAC names are used ubiquitously in scholarly articles, patents, and educational materials. %
In contrast to other linear molecular representations like SMILES and its derivatives, where single tokens mostly refer to individual atoms and bonds, tokens in IUPAC names generally have a rich semantic meaning.
For example, the token ``ic acid'' denotes a carboxylic acid, which is a common functional group that has well-known physical and chemical properties; there are many known chemical reactions that either start with or produce carboxylic acids.
Other tokens denote additional functional groups (e.g. ``imide,'' ``imine,'' ``al,'' ``one''), locants (e.g. ``1,'' ``2,'' ``N''), which indicate connectivity, alkanes (e.g. ``meth,'' ``eth,'' ``prop''), which denote the lengths of carbon chains, polycyclic rings (e.g. ``naphthalene,'' ``anthracene''), stereochemistry markers (``R,'' ``S''), and multipliers (e.g. ``di,'' ``tri''), which concisely represent duplicated and symmetric structures.
Figure \ref{fig:naming} shows the relationships between IUPAC names, graph representations, and SMILES.

For molecular optimization, C5T5 supports qualitatively different molecular edits compared to graph- and SMILES-based methods by virtue of its use of IUPAC names.
In particular, editing a locant token corresponds to moving a functional group along a carbon backbone or changing the connectivity of a fused ring system.
And editing a multiplier token corresponds to creating or eliminating duplicated and symmetric structures.
For example, changing ``ethylbenzene`` to ``hexaethylbenzene'' replicates the ethyl structure around the entire benzene ring with a single token edit.
These sorts of modifications require much more extensive editing for SMILES- and graph-based methods.\footnote{Moving the attachment point of a functional group requires only a small edit of a graph (i.e. changing one bond), but most graph-based molecular generation methods sequentially generate one node at a time, followed by any bonds that connect the new node to the molecular graph so far. Moving a side chain therefore requires removing the entire chain and regenerating it node by node.}

We argue that IUPAC names are especially attractive for molecular optimization, since the process requires interaction between the algorithm and a domain expert, so interpretability is paramount.
Compared to graph- or SMILES-based models, C5T5 makes predictions that can be traced back to moieties and functional groups that domain experts are more likely to understand, trust, and know how to synthesize than arbitrary collections of atoms and bonds.

\begin{figure}
    \centering
    \includegraphics[width=\textwidth]{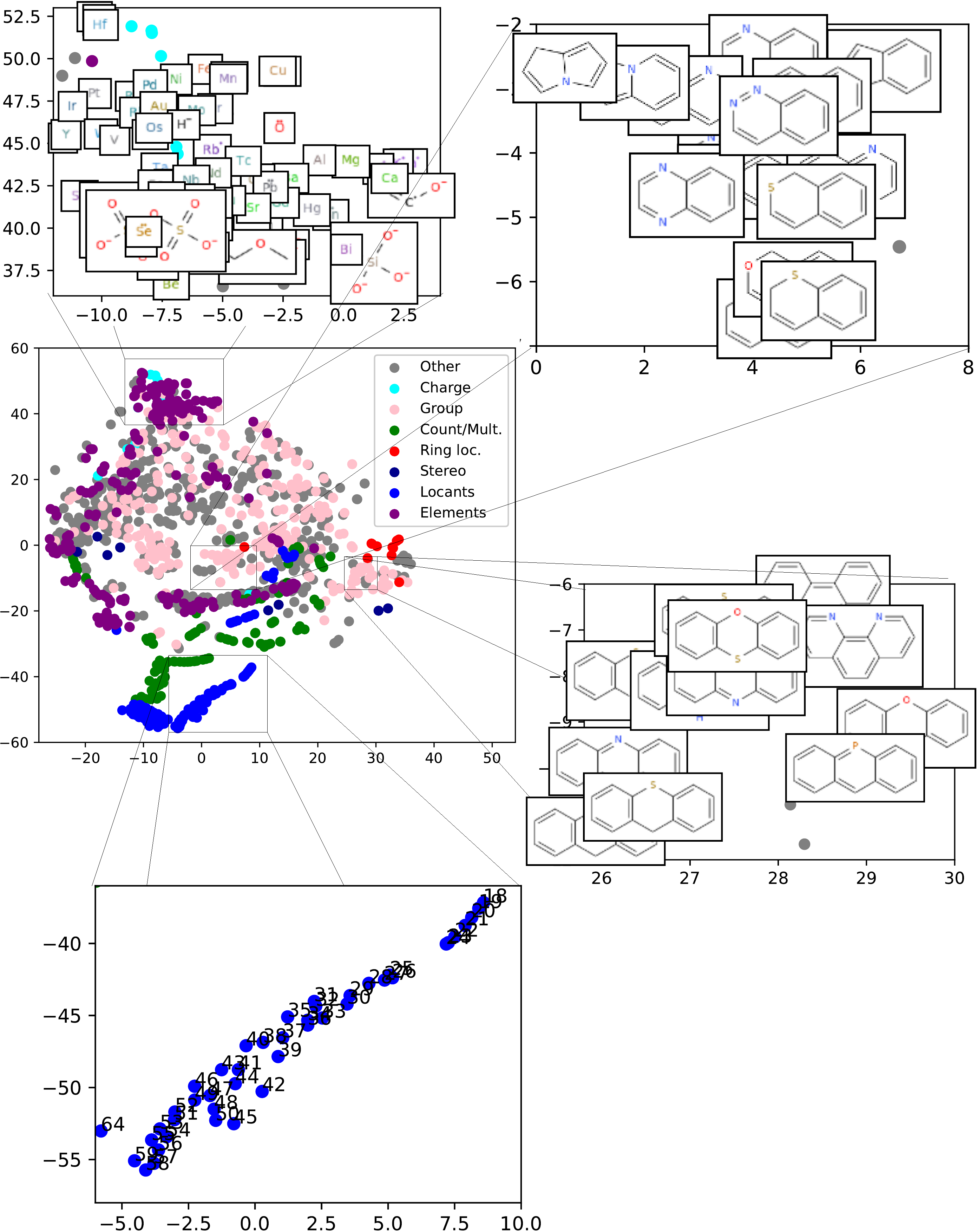}
    \caption{\textbf{T-SNE visualization of the word2vec embedding space.} ``Charge'': tokens that indicate formal charge. ``Group'': functional groups and moieties. ``Count/Mult'': multipliers. ``Ring loc.'': fused-ring locants. ``Stereo'': stereochemistry markers. ``Locants'': simple locants. ``Elements'': single-atom tokens. As shown, the 2D location of tokens carries high semantic meaning; for example, locants are not only collocated, but are approximately in order.}
    \label{fig:word2vec}
\end{figure}

In addition to improved interpretability, we argue that using IUPAC names has advantages purely from the standpoint of modeling data, since moving from SMILES to IUPAC names is akin to moving from a character-based to a word-based sequence model.
Modeling at this higher level of abstraction enables the network to direct more of its capacity to structure at the relevant semantic level, instead of relearning lower-level details like the specific atomic composition of functional groups.
In this vein, we demonstrate the potential of IUPAC names by learning word2vec representations of IUPAC name tokens \cite{word2vec}, drawn from a list of over 100 million names in the PubChem repository \citep{kim2016pubchem} and tokenized using a list of tokens in OPSIN---an open-source IUPAC Name parser library (MIT License) \cite{opsin}.
For example, as shown in Figure \ref{fig:naming}, the chemical ``2-acetyloxybenzoic acid'' gets tokenized to [``2'', ``-'', ``acet'', ``yl'', ``oxy'', ``benzo'', ``ic acid''].
As with natural language modeling, we find that the embedding space learned by word2vec encodes the semantic meaning of the tokens, as shown in Figure \ref{fig:word2vec}.
Different classes of tokens tend to be clustered together, and similar tokens within clusters are located nearby.
For example, aromatic compounds with two rings are clearly separated from those with three, locants are ordered roughly correctly from 1 to 100, and multiplier tokens are also roughly in order (zoom not shown).
Following \citet{word2vec}, we also find that simple arithmetic operations in the embedding vector space correspond to semantic analogies between tokens.
For example, the nearest neighbor of ``phosphonous acid'' - ``nitrous acid'' + ``nitroso'' is the embedding for ``phosphoroso.''\footnote{ignoring the embedding for ``nitroso''}
The nearest neighbor of ``diphosphate'' - ``disulfate'' + ``sulfate'' is ``phosphate.''
Likewise for ``selenate'' - ``tellurate'' + ``tellurite'' being closest to ``selenite.''

\subsection{Conditional Language Modeling with Transformers}
\label{sec:method_t5}

We now present the C5T5 objective, which trains a model to alter the properties of a molecule through localized edits. Importantly, this behavior does not require training on human-defined edit pairs, a strategy limited by either a fixed set of hand-specified chemical alterations or an expensive experimentally-derived training set. Instead, this editing behavior emerges as a zero-shot side-effect of our conditional language modeling objective, requiring only a simple forward pass using our pretrained model without additional gradient updates.

Given a property value $P$, a fragment of a molecule $F$, and the rest of the molecule $C$ (the context), we wish to learn the conditional distribution $P(F | C, P)$. Then, for a new molecule, one could alter the molecule towards a desired property value by redacting the original $F$, changing $P$ to $P'$ and sampling a new $F' \sim P(F | C, P')$. Intuitively, this asks our model what kinds of molecular fragments the model would expect given the context and the new property value.

To learn this conditional distribution, we propose a conditional generalization of the infilling objective used to train T5 \citep{t5} and ILM \citep{donahue2020ilm}. This process consists of several steps, also illustrated in Figure \ref{fig:objective}:

\begin{enumerate}
    \item Replacing random spans of the tokenized IUPAC name with sentinel tokens.
    \item Prepending the resulting sequence with a token indicating the original molecule's computed property value. To obtain these property value tokens, we discretize the distribution of property values into three buckets, specified in Table \ref{tab:property-ranges}.  
    \item Training the model as in T5 to produce the sequence of redacted tokens, prepended by their corresponding sentinel tokens.
\end{enumerate}

\begin{table}[h]
    \centering
    \begin{tabular}{lccc} \toprule
        Property & \texttt{<low>} & \texttt{<med>} & \texttt{<high>} \\
        \midrule
        Octanol-water partition coeff. (logP) & $(-\infty, -0.4)$ & $(-0.4, 5.6)$ & $(5.6, \infty)$\\
        Octanol-water distribution coeff. (logD) & $(-\infty, -0.4)$ & $(-0.4, 5.6)$ & $(5.6, \infty)$\\
        Polar surface area (PSA) & $(0, 90)$ & $(90, 140)$ & $(140, \infty)$ \\
        Refractivity & $(0, 40)$ & $(40, 130)$ & $(130, \infty)$ \\
        \bottomrule\\
    \end{tabular}
    \caption{\textbf{Numerical ranges across properties for each property value token.} Cutoffs for logP, PSA, and refractivity were chosen as common thresholds for druglikeness screening following \cite{ghose1999knowledge, veber2002molecular, hitchcock2006structure}. We use the same cutoff for logD as logP.}
    \label{tab:property-ranges}
\end{table}

This conditional infilling objective incentivizes the model to learn the relationship between the computed property value and the missing tokens. To make a localized edit to a molecule, we then replace the desired fragments with sentinel tokens, change the property value token, and sample autoregressively from the predictive distribution. Thus, our approach hybridizes the flexible editing capabilities of ILM \citep{donahue2020ilm} with the controllability of CTRL \citep{Keskar2019CTRLAC}.
See Appendix \ref{app:experiment_details} for experimental details. Code is available at \url{https://github.com/dhroth/c5t5}.

\section{Results}
\label{sec:results}

To demonstrate the promise of combining IUPAC names with conditional modeling using T5, we explore several molecular optimization tasks relevant to drug discovery.
Specifically, we train C5T5 to make localized changes that affect the octanol-water partition and distribution coefficients (logP, logD), polar surface area (PSA), and refractivity---four properties commonly used to estimate bioavailability of a candidate drug \cite{ghose1999knowledge, veber2002molecular, bhal2007rule}.
logP and logD measure the ratio of a compound's solubility in octanol, a lipid-like structure, and water; drugs need to be somewhat soluble in both to be orally absorbed.
PSA and refractivity both relate to charge separation within the molecule.
As shown in Appendix \ref{app:novelty}, C5T5 generates mostly valid and novel molecules, with values of logP that lie outside of the range of  a ``best in dataset'' baseline for targeted modifications.

\subsection{C5T5 Successfully Modifies Properties}
\label{subsec:results-modifies}
\begin{figure}
    \centering
    \begin{subfigure}{0.5\textwidth}
    \includegraphics[width=0.98\textwidth]{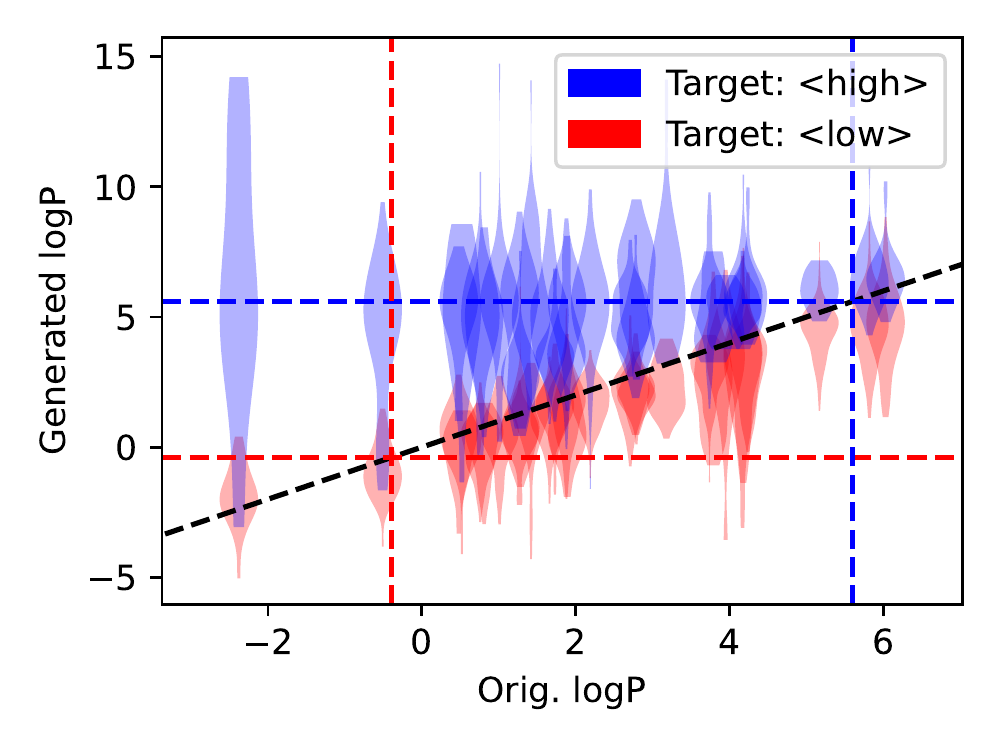}
    \end{subfigure}%
    \begin{subfigure}{0.5\textwidth}
    \includegraphics[width=0.98\textwidth]{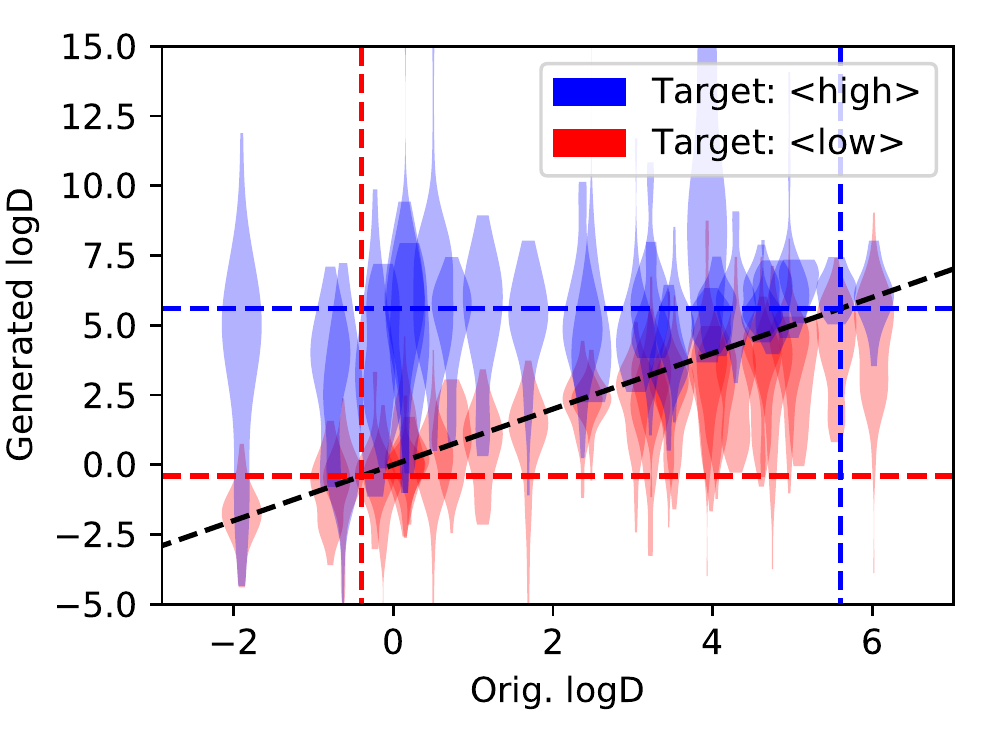}
    \end{subfigure}
    \begin{subfigure}{0.5\textwidth}
    \includegraphics[width=0.98\textwidth]{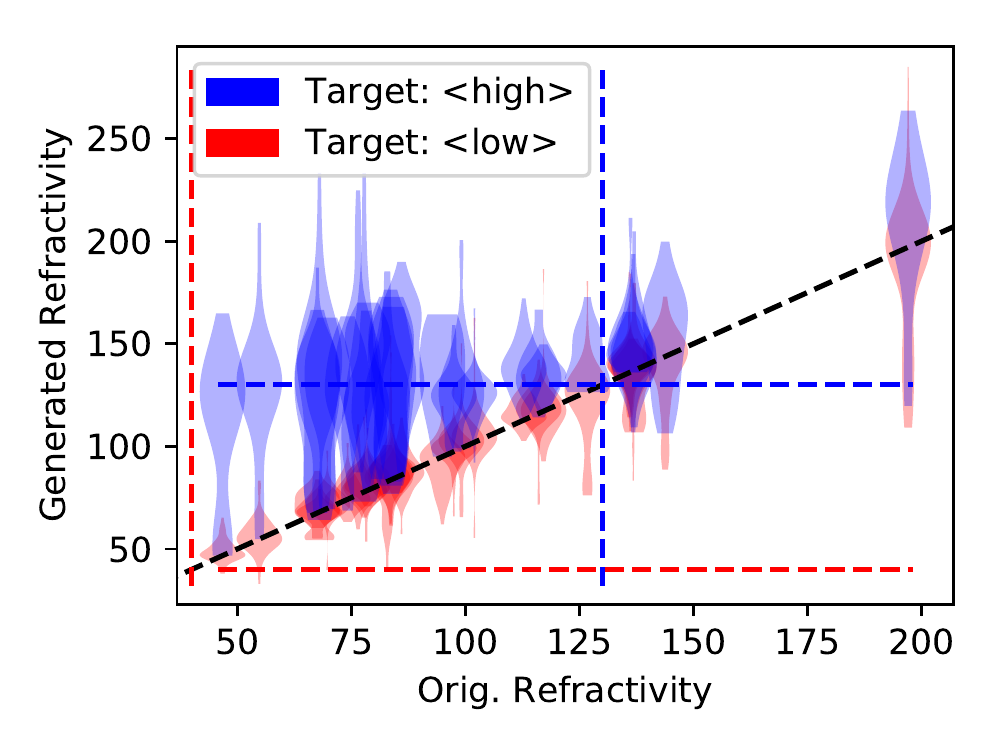}
    \end{subfigure}%
    \begin{subfigure}{0.5\textwidth}
    \includegraphics[width=0.98\textwidth]{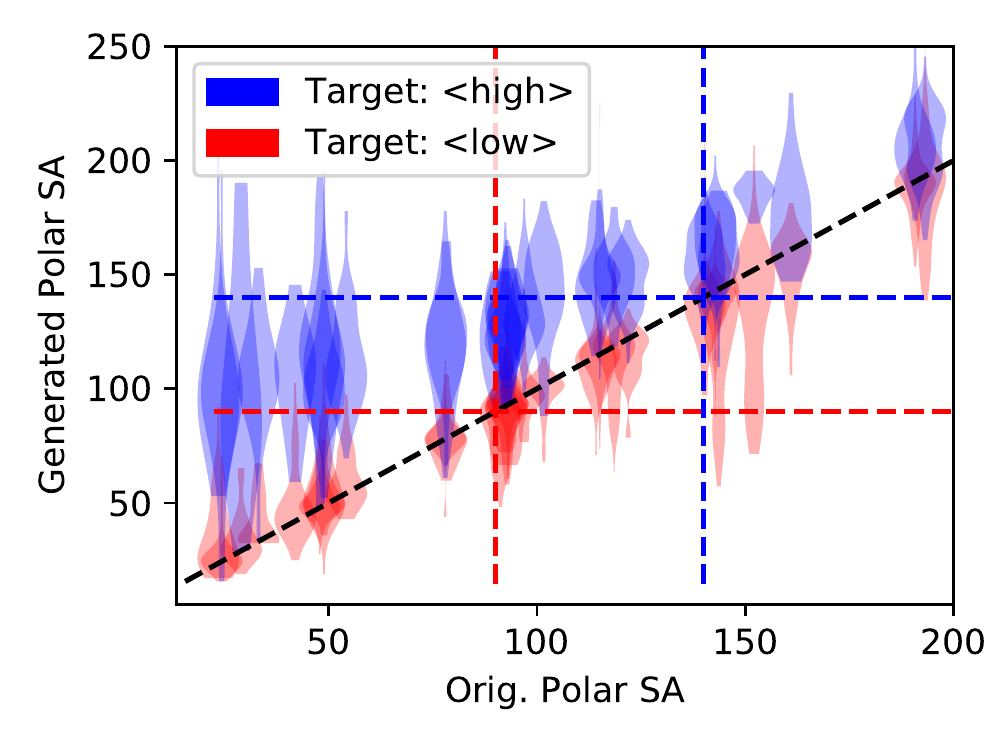}
    \end{subfigure}
    \caption{\textbf{Calculated property values of optimized molecules vs. original.} Values computed for 30 randomly chosen starting molecules. Top left: octanol-water partition coefficient. Top right: octanol-water distribution coefficient at pH of 7. Bottom left: molar refractivity. Bottom right: polar surface area. Blue violins show the distribution of generated molecule properties when the model was asked to complete molecules to achieve a \texttt{<high>} property value, and red for \texttt{<low>}. Cutoffs between \texttt{<high>} and \texttt{<med>} and between \texttt{<med>} and \texttt{<low>} are shown as dashed blue and red lines, respectively. The black dashed line is y=x.}
    \label{fig:gen_vs_orig}
\end{figure}

First, we demonstrate that the localized changes proposed by C5T5 do in fact control the property value as desired.
C5T5 allows domain experts to choose where to make changes to a molecule based on their intuition and particular application: the user masks out tokens in the IUPAC name that can be modified, and the model fills in the masked spans in a way that changes the property value as directed.
There is no canonical way to choose particular tokens to mask for evaluation purposes, so we simply choose a number of starting molecules randomly from PubChem \cite{kim2016pubchem}, and then we iteratively mask all length-one to length-five spans (to match the training distribution) and run inference.
In some instances, there is only one way to fill in the mask that yields a valid IUPAC name, regardless of the desired property value (e.g. a single-token mask that removes a parenthesis, a comma between two locants, or a hyphen after a locant).
This would not be an issue in real usage, so in our evaluation we simply ignore cases when the model generates the starting molecule.
We also experiment with masking multiple spans per molecule during inference, as is done during training.
This is useful in practice when there are multiple areas of the molecule that can be changed in tandem to achieve the desired property value, but in our evaluation we observe qualitatively similar results when masking only a single span, so for computational efficiency we limit ourselves to single spans.
We expect multi-span masks to be much more important when optimizing for more complex properties.

Figure \ref{fig:gen_vs_orig} shows that C5T5 successfully generates molecules with higher property values when passed <high>, and with lower property values when passed <low>.
The model is much more successful at raising property values than lowering them, especially for refractivity and polar surface area.
For both of these properties, increasing the property value is straightforward: just add polar groups to replace whatever tokens were masked.
In contrast, lowering these properties is only possible when the mask coincides with a polar group, in which case the model must find a non-polar substitute while still maintaining the molecule's validity.
Even if unsuccessful at lowering these two property values on average, C5T5 can still be used in this case to suggest a number of candidate edits, and the one with the lowest property value can be selected using a property prediction model.
This is an improvement over high-throughput screening and untargeted machine-learning methods for molecular optimization, since it isn't restricted to a predefined library of candidate molecules, and it still allows the user to choose particular parts of the molecule to modify.

\begin{table}[h]
    \caption{\textbf{Tokens most preferentially added when C5T5 is asked to make modifications resulting high vs. low logP values.} Multipliers compare the actual rate of adding tokens compared to the expected number if the model drew randomly from the data distribution independent of property value. \textcolor{blue}{Blue tokens} are hydrocarbons (i.e. lipophilic groups). \textcolor{red}{Red tokens} contain hydrogen bonding donors or acceptors (i.e. hydrophilic groups)}
    \label{tab:common_changes}
    \centering
    \begin{tabular}{lrlr}
        \\\toprule
         \multicolumn{2}{c}{\textbf{Target: \texttt{<high>}}} & {\textbf{Target: \texttt{<low>}}} \\
         \midrule
        \textcolor{blue}{trityl} & 77.0x     & \textcolor{red}{phospho} & 48.1x\\ 
        \textcolor{blue}{pentadecyl} & 20.2x & \textcolor{red}{phosphonato} & 44.7x\\ 
        Z & 17.7x                            & \textcolor{red}{sulfinam} & 41.2x\\ 
        \textcolor{blue}{perylen} & 11.8x    & \textcolor{red}{hydrazon} & 34.3x\\ 
        \textcolor{blue}{undecyl} & 8.1x     & \textcolor{red}{sulfinato} & 26.3x\\ 
        \textcolor{blue}{heptadecyl} & 7.9x  & Z & 17.9x\\ 
        ylium & 7.6x                         & \textcolor{red}{oxonium} & 10.3x\\ 
        \textcolor{red}{isoindolo} & 5.9x    & \textcolor{red}{amoyl} & 9.3x\\ 
        bH & 5.9x                            & \textcolor{red}{carbamic acid} & 8.6x\\
        iod & 5.8x                           & \textcolor{red}{sulfin} & 6.9x \\
        \bottomrule
    \end{tabular}
\end{table}

\begin{figure}
    \centering
    \includegraphics[width=0.9\textwidth]{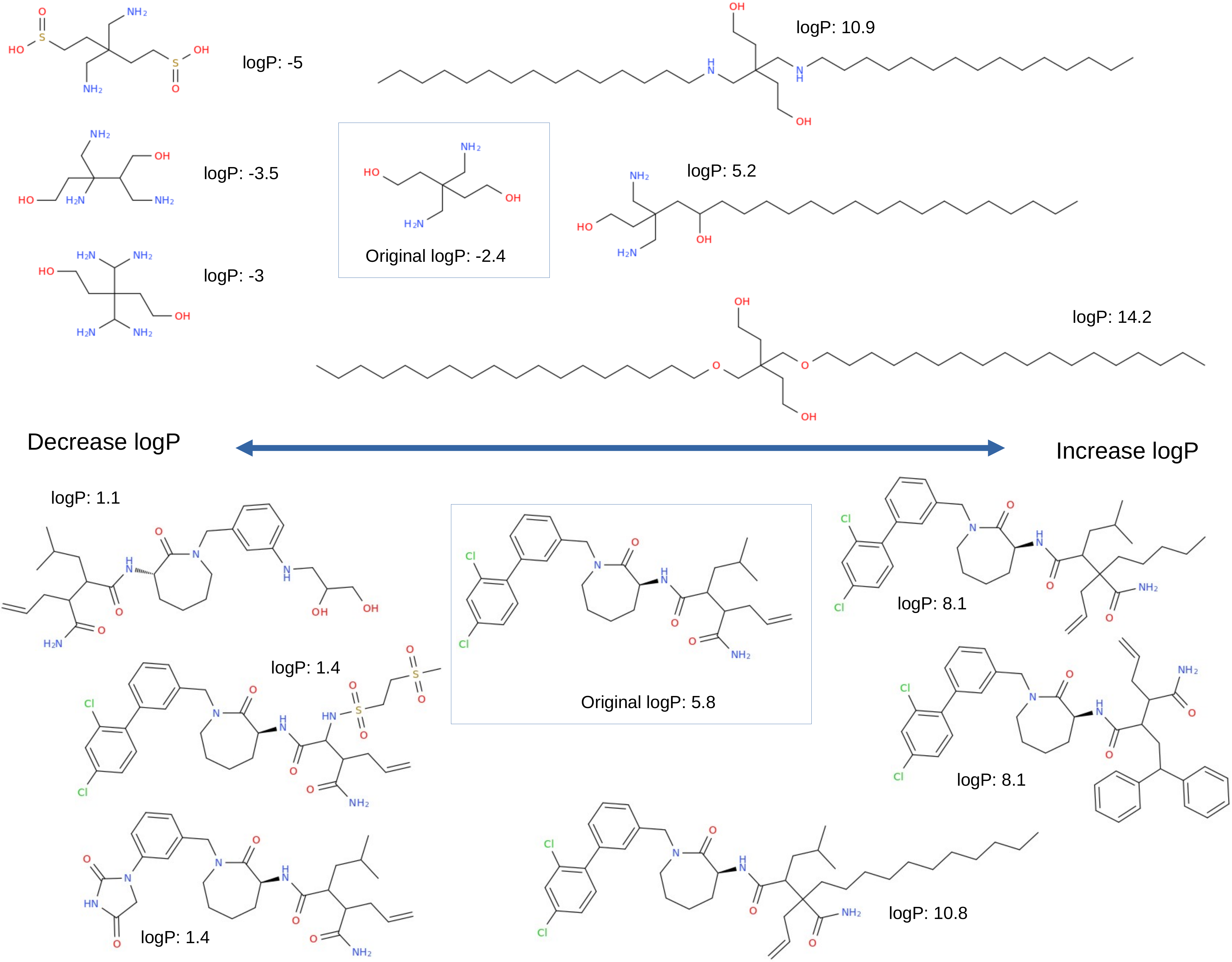}
    \caption{\textbf{Visualizations of base and logP-optimized molecules}. Two molecules from the logP plot in Figure \ref{fig:gen_vs_orig}, with three molecules after optimizing with C5T5 for each of <low> and <high> logP.}
    \label{fig:viz_logp}
\end{figure}

\subsection{Modified Tokens are Chemically Intuitive}
\label{subsec:results-intuitive}
One main advantage of C5T5 is that the changes it suggests are in the intuitive language of IUPAC names, rather than suggesting arbitrary combinations of atoms.
Table \ref{tab:common_changes} shows the tokens that the model most preferentially adds to a molecule when asked to produce low vs. high logP.
Unsurprisingly, the most common tokens added when increasing logP are generally long carbon chains (pentadecyl, undecyl, heptadecyl) and other hydrocarbons (trityl, perylen).
Conversely, when the model is asked to produce low-logP modifications, hydrophilic groups are added.
LogP is a simple metric, and by proposing molecular edits for logP that are obvious and easily understandable to domain experts, we expect users to gain confidence that C5T5 will suggest reasonable edits for more complex properties.

To further investigate the types of optimizations C5T5 suggests, Figure \ref{fig:viz_logp} visualizes two of the starting molecules from Figure \ref{fig:gen_vs_orig} with logP values of $-2.4$ and $5.8$.
For each molecule, we mask spans as usual and generate after prepending with <low> (molecules on the left) and <high> (molecules on the right).
The IUPAC name of the top starting molecule is ``3,3-bis(aminomethyl)pentane-1,5-diol,'' where ``bis'' signifies that the CNH2 group should be duplicated, and ``diol'' means duplicate OH groups at the ends.
By virtue of the IUPAC name encoding symmetry, C5T5 is easily able to generate similarly symmetric molecules.
For example, the molecule in the top left is ``3,3-bis(aminomethyl)pentane-1,5-disulfinic acid,'' where the ``ol'' has been replaced with ``sulfinic acid.''
Although symmetry is often highly desired, C5T5 is not limited to generating symmetric molecules.
For example, the middle molecule on the right is ``3,3-bis(aminomethyl)-1-heptadecylpentane-1,5-diol''---C5T5 added an additional carbon chain non-symmetrically at the end of the pentane.\footnote{Although this is not a preferred IUPAC name, it is still unambiguous, and therefore valid and parseable.}

Sometimes C5T5 generates valid but unstable compounds.
For example, the neighboring NH$_2$ groups in the bottom left molecule in the top half of Figure \ref{fig:viz_logp} are unstable, and would turn into aldehydes in an aqueous solution.
All machine learning methods are susceptible to this sort of mistake, underscoring the importance of the type of human-in-the-loop optimization that C5T5 enables.

\section{Discussion and Conclusion}
\label{sec:discussion}

We propose C5T5, a simple and effective zero-shot method for targeted control of molecular properties with transformers.
Unlike prior approaches that make user-targeted modifications, our method requires no database of paired edits; instead, it simply trains in a self-supervised fashion on a large dataset of molecules and coarse estimates of their molecular property values.
Core to our method is the use of IUPAC names as a data representation, which captures molecular structure at an appropriate level of abstraction and enables an intuitive editing interface for domain experts.
C5T5 successfully rediscovers chemically-intuitive strategies for altering four drug-related properties in molecules, a notable feat given the absence of any human demonstration of these editing strategies. 

Our work also has several limitations.
The select-and-replace interface provided by the infilling objective may not always match the needs or preferred design process of biochemists who might wish to use it.
The interface also only suggests how to fill in missing parts of a molecule, and it relies on domain expertise or enumeration to decide which parts of the molecule should be changed to begin with.
In addition, we explore only a coarse-grained bucketing of property values, leaving a more fine-grained treatment for future work.
IUPAC names might also be too limiting in cases where a user wants to edit a subgraph of a molecule that does not correspond neatly to a small number of IUPAC tokens.
Finally, although there are many potential positive impacts to better drug design, including safer and more effective medicines to prolong the length and quality of people's lives, the generality of this method means that bad actors could optimize molecules towards harmful properties as well.

Future work will investigate using C5T5 with more molecular properties, such as the power conversion efficiency of solar cells.
We also leave to future work extending C5T5 to jointly model multiple properties, and adding a more flexible editing interface.
\section{Acknowledgements}
ChemAxon's calculator was used to compute IUPAC names, chemical descriptors, and molecular properties (version 20.17.0, https://www.chemaxon.com).

We thank ACD/Labs for computing IUPAC names for several datasets.

This material is based upon work supported by the National Science Foundation Graduate Research Fellowship Program under Grant No. DGE 1752814. Any opinions, findings, and conclusions or recommendations expressed in this material are those of the author(s) and do not necessarily reflect the views of the National Science Foundation.

AT is supported by an OpenPhil AI fellowship.

In addition to NSF CISE Expeditions Award CCF-1730628, this research is supported by gifts from Amazon Web Services, Ant Group, Ericsson, Facebook, Futurewei, Google, Intel, Microsoft, Nvidia, Scotiabank, Splunk and VMware.

\newpage

\bibliography{refs}

\begin{thebibliography}{58}
\providecommand{\natexlab}[1]{#1}
\providecommand{\url}[1]{\texttt{#1}}
\expandafter\ifx\csname urlstyle\endcsname\relax
  \providecommand{\doi}[1]{doi: #1}\else
  \providecommand{\doi}{doi: \begingroup \urlstyle{rm}\Url}\fi

\bibitem[Reymond et~al.(2012)Reymond, Ruddigkeit, Blum, and van
  Deursen]{enumerate-wircms12}
Jean-Louis Reymond, Lars Ruddigkeit, Lorenz Blum, and Ruud van Deursen.
\newblock The enumeration of chemical space.
\newblock \emph{Wiley Interdisciplinary Reviews: Computational Molecular
  Science}, 2\penalty0 (5):\penalty0 717--733, 2012.

\bibitem[Hughes et~al.(2011)Hughes, Rees, Kalindjian, and
  Philpott]{principles-bjp11}
James~P Hughes, Stephen Rees, S~Barrett Kalindjian, and Karen~L Philpott.
\newblock Principles of early drug discovery.
\newblock \emph{British journal of pharmacology}, 162\penalty0 (6):\penalty0
  1239--1249, 2011.

\bibitem[Raffel et~al.(2019)Raffel, Shazeer, Roberts, Lee, Narang, Matena,
  Zhou, Li, and Liu]{t5}
Colin Raffel, Noam Shazeer, Adam Roberts, Katherine Lee, Sharan Narang, Michael
  Matena, Yanqi Zhou, Wei Li, and Peter~J Liu.
\newblock Exploring the limits of transfer learning with a unified text-to-text
  transformer.
\newblock \emph{arXiv preprint arXiv:1910.10683}, 2019.

\bibitem[Weininger(1988)]{smiles}
David Weininger.
\newblock Smiles, a chemical language and information system. 1. introduction
  to methodology and encoding rules.
\newblock \emph{Journal of chemical information and computer sciences},
  28\penalty0 (1):\penalty0 31--36, 1988.

\bibitem[Duvenaud et~al.(2015)Duvenaud, Maclaurin, Aguilera-Iparraguirre,
  G\'{o}mez-Bombarelli, Hirzel, Aspuru-Guzik, and
  Adams]{molfingerprints-neurips15}
David Duvenaud, Dougal Maclaurin, Jorge Aguilera-Iparraguirre, Rafael
  G\'{o}mez-Bombarelli, Timothy Hirzel, Al\'{a}n Aspuru-Guzik, and Ryan~P.
  Adams.
\newblock Convolutional networks on graphs for learning molecular fingerprints.
\newblock In \emph{Proceedings of the 28th International Conference on Neural
  Information Processing Systems - Volume 2}, NIPS'15, page 2224–2232,
  Cambridge, MA, USA, 2015. MIT Press.

\bibitem[Ghosh et~al.(2017)Ghosh, Chollet, Laksana, Morency, and
  Scherer]{ghosh2017affect}
Sayan Ghosh, Mathieu Chollet, Eugene Laksana, Louis-Philippe Morency, and
  Stefan Scherer.
\newblock Affect-lm: A neural language model for customizable affective text
  generation.
\newblock \emph{arXiv preprint arXiv:1704.06851}, 2017.

\bibitem[Ficler and Goldberg(2017)]{Ficler2017ControllingLS}
Jessica Ficler and Y.~Goldberg.
\newblock Controlling linguistic style aspects in neural language generation.
\newblock \emph{ArXiv}, abs/1707.02633, 2017.

\bibitem[Niu and Bansal(2018)]{niu2018polite}
Tong Niu and Mohit Bansal.
\newblock Polite dialogue generation without parallel data.
\newblock \emph{Transactions of the Association for Computational Linguistics},
  6:\penalty0 373--389, 2018.

\bibitem[Keskar et~al.(2019)Keskar, McCann, Varshney, Xiong, and
  Socher]{Keskar2019CTRLAC}
N.~Keskar, Bryan McCann, L.~Varshney, Caiming Xiong, and R.~Socher.
\newblock Ctrl: A conditional transformer language model for controllable
  generation.
\newblock \emph{ArXiv}, abs/1909.05858, 2019.

\bibitem[Grisoni et~al.(2020)Grisoni, Moret, Lingwood, and
  Schneider]{bimodal-lstm-jcim20}
Francesca Grisoni, Michael Moret, Robin Lingwood, and Gisbert Schneider.
\newblock Bidirectional molecule generation with recurrent neural networks.
\newblock \emph{Journal of chemical information and modeling}, 60\penalty0
  (3):\penalty0 1175--1183, 2020.

\bibitem[Guimaraes et~al.(2017)Guimaraes, Sanchez-Lengeling, Outeiral, Farias,
  and Aspuru-Guzik]{organ-arxiv17}
Gabriel~Lima Guimaraes, Benjamin Sanchez-Lengeling, Carlos Outeiral, Pedro
  Luis~Cunha Farias, and Al{\'a}n Aspuru-Guzik.
\newblock Objective-reinforced generative adversarial networks (organ) for
  sequence generation models.
\newblock \emph{arXiv preprint arXiv:1705.10843}, 2017.

\bibitem[Sanchez-Lengeling et~al.()Sanchez-Lengeling, Outeiral, Guimaraes, and
  Aspuru-Guzik]{sanchez2017optimizing}
Benjamin Sanchez-Lengeling, Carlos Outeiral, Gabriel~L Guimaraes, and Alan
  Aspuru-Guzik.
\newblock Optimizing distributions over molecular space. an
  objective-reinforced generative adversarial network for inverse-design
  chemistry (organic).

\bibitem[De~Cao and Kipf(2018)]{molgan-icmlw18}
Nicola De~Cao and Thomas Kipf.
\newblock {MolGAN: An implicit generative model for small molecular graphs}.
\newblock \emph{ICML 2018 workshop on Theoretical Foundations and Applications
  of Deep Generative Models}, 2018.

\bibitem[Kajino(2019)]{molhyper-vae-icml19}
Hiroshi Kajino.
\newblock Molecular hypergraph grammar with its application to molecular
  optimization.
\newblock In \emph{International Conference on Machine Learning}, pages
  3183--3191. PMLR, 2019.

\bibitem[He et~al.(2021{\natexlab{a}})He, You, Sandstr{\"o}m, Nittinger,
  Bjerrum, Tyrchan, Czechtizky, and Engkvist]{mmp-transformer-jcheminfo21}
Jiazhen He, Huifang You, Emil Sandstr{\"o}m, Eva Nittinger, Esben~Jannik
  Bjerrum, Christian Tyrchan, Werngard Czechtizky, and Ola Engkvist.
\newblock Molecular optimization by capturing chemist’s intuition using deep
  neural networks.
\newblock \emph{Journal of cheminformatics}, 13\penalty0 (1):\penalty0 1--17,
  2021{\natexlab{a}}.

\bibitem[Jin et~al.(2019)Jin, Yang, Barzilay, and Jaakkola]{vjtnn-iclr19}
Wengong Jin, Kevin Yang, Regina Barzilay, and Tommi Jaakkola.
\newblock Learning multimodal graph-to-graph translation for molecule
  optimization.
\newblock In \emph{International Conference on Learning Representations}, 2019.
\newblock URL \url{https://openreview.net/forum?id=B1xJAsA5F7}.

\bibitem[Shin et~al.(2021)Shin, Park, Bak, and Ho]{cmg-transformer-chil21}
Bonggun Shin, Sungsoo Park, JinYeong Bak, and Joyce~C Ho.
\newblock Controlled molecule generator for optimizing multiple chemical
  properties.
\newblock In \emph{Proceedings of the Conference on Health, Inference, and
  Learning}, pages 146--153, 2021.

\bibitem[Yang et~al.(2020)Yang, Jin, Swanson, Barzilay, and
  Jaakkola]{yang2020improving}
Kevin Yang, Wengong Jin, Kyle Swanson, Regina Barzilay, and Tommi Jaakkola.
\newblock Improving molecular design by stochastic iterative target
  augmentation.
\newblock In \emph{International Conference on Machine Learning}, pages
  10716--10726. PMLR, 2020.

\bibitem[Kotsias et~al.(2020)Kotsias, Ar{\'u}s-Pous, Chen, Engkvist, Tyrchan,
  and Bjerrum]{crnn-nature-mi20}
Panagiotis-Christos Kotsias, Josep Ar{\'u}s-Pous, Hongming Chen, Ola Engkvist,
  Christian Tyrchan, and Esben~Jannik Bjerrum.
\newblock Direct steering of de novo molecular generation with descriptor
  conditional recurrent neural networks.
\newblock \emph{Nature Machine Intelligence}, 2\penalty0 (5):\penalty0
  254--265, 2020.

\bibitem[Gómez-Bombarelli et~al.(2018)Gómez-Bombarelli, Wei, Duvenaud,
  Hernández-Lobato, Sánchez-Lengeling, Sheberla, Aguilera-Iparraguirre,
  Hirzel, Adams, and Aspuru-Guzik]{bombarelli-acs-cs18}
Rafael Gómez-Bombarelli, Jennifer~N. Wei, David Duvenaud, José~Miguel
  Hernández-Lobato, Benjamín Sánchez-Lengeling, Dennis Sheberla, Jorge
  Aguilera-Iparraguirre, Timothy~D. Hirzel, Ryan~P. Adams, and Alán
  Aspuru-Guzik.
\newblock Automatic chemical design using a data-driven continuous
  representation of molecules.
\newblock \emph{ACS Central Science}, 4\penalty0 (2):\penalty0 268--276, 2018.
\newblock \doi{10.1021/acscentsci.7b00572}.
\newblock URL \url{https://doi.org/10.1021/acscentsci.7b00572}.
\newblock PMID: 29532027.

\bibitem[Lim et~al.(2018)Lim, Ryu, Kim, and Kim]{condvae-jcheminfo18}
Jaechang Lim, Seongok Ryu, Jin~Woo Kim, and Woo~Youn Kim.
\newblock Molecular generative model based on conditional variational
  autoencoder for de novo molecular design.
\newblock \emph{Journal of cheminformatics}, 10\penalty0 (1):\penalty0 1--9,
  2018.

\bibitem[Dollar et~al.(2021)Dollar, Joshi, Beck, and
  Pfaendtner]{vae-transformer-cs21}
Orion Dollar, Nisarg Joshi, David Beck, and Jim Pfaendtner.
\newblock Attention-based generative models for de novo molecular design.
\newblock \emph{Chemical Science}, 2021.

\bibitem[Liu et~al.(2018)Liu, Allamanis, Brockschmidt, and
  Gaunt]{cgvae-neurips18}
Qi~Liu, Miltiadis Allamanis, Marc Brockschmidt, and Alexander~L. Gaunt.
\newblock Constrained graph variational autoencoders for molecule design.
\newblock In \emph{Proceedings of the 32nd International Conference on Neural
  Information Processing Systems}, NIPS'18, page 7806–7815, Red Hook, NY,
  USA, 2018. Curran Associates Inc.

\bibitem[Jin et~al.(2018)Jin, Barzilay, and Jaakkola]{jtvae-icml18}
Wengong Jin, Regina Barzilay, and Tommi Jaakkola.
\newblock Junction tree variational autoencoder for molecular graph generation.
\newblock In \emph{International Conference on Machine Learning}, pages
  2323--2332. PMLR, 2018.

\bibitem[Maziarka et~al.(2020)Maziarka, Pocha, Kaczmarczyk, Rataj, Danel, and
  Warcho{\l}]{cyclegan-jchi20}
{\L}ukasz Maziarka, Agnieszka Pocha, Jan Kaczmarczyk, Krzysztof Rataj, Tomasz
  Danel, and Micha{\l} Warcho{\l}.
\newblock Mol-cyclegan: a generative model for molecular optimization.
\newblock \emph{Journal of Cheminformatics}, 12\penalty0 (1):\penalty0 1--18,
  2020.

\bibitem[Olivecrona et~al.(2017)Olivecrona, Blaschke, Engkvist, and
  Chen]{olivecrona2017molecular}
Marcus Olivecrona, Thomas Blaschke, Ola Engkvist, and Hongming Chen.
\newblock Molecular de-novo design through deep reinforcement learning.
\newblock \emph{Journal of cheminformatics}, 9\penalty0 (1):\penalty0 1--14,
  2017.

\bibitem[Bagal et~al.(2021)Bagal, Aggarwal, Vinod, and
  Priyakumar]{liggpt-chemrxiv21}
Viraj Bagal, Rishal Aggarwal, PK~Vinod, and U~Deva Priyakumar.
\newblock Liggpt: Molecular generation using a transformer-decoder model.
\newblock 2021.

\bibitem[You et~al.(2018)You, Liu, Ying, Pande, and Leskovec]{gcpn-neurips18}
Jiaxuan You, Bowen Liu, Rex Ying, Vijay Pande, and Jure Leskovec.
\newblock Graph convolutional policy network for goal-directed molecular graph
  generation.
\newblock In \emph{Proceedings of the 32nd International Conference on Neural
  Information Processing Systems}, NIPS'18, page 6412–6422, Red Hook, NY,
  USA, 2018. Curran Associates Inc.

\bibitem[Shi* et~al.(2020)Shi*, Xu*, Zhu, Zhang, Zhang, and
  Tang]{graphaf-iclr20}
Chence Shi*, Minkai Xu*, Zhaocheng Zhu, Weinan Zhang, Ming Zhang, and Jian
  Tang.
\newblock Graphaf: a flow-based autoregressive model for molecular graph
  generation.
\newblock In \emph{International Conference on Learning Representations}, 2020.
\newblock URL \url{https://openreview.net/forum?id=S1esMkHYPr}.

\bibitem[Li et~al.(2019)Li, Hu, Wang, Zhou, Zhang, and
  Liu]{deepscaffold-jcim19}
Yibo Li, Jianxing Hu, Yanxing Wang, Jielong Zhou, Liangren Zhang, and Zhenming
  Liu.
\newblock Deepscaffold: A comprehensive tool for scaffold-based de novo drug
  discovery using deep learning.
\newblock \emph{Journal of chemical information and modeling}, 60\penalty0
  (1):\penalty0 77--91, 2019.

\bibitem[Lim et~al.(2020)Lim, Hwang, Moon, Kim, and
  Kim]{scaffold-based-vae-rsc-cs20}
Jaechang Lim, Sang-Yeon Hwang, Seokhyun Moon, Seungsu Kim, and Woo~Youn Kim.
\newblock Scaffold-based molecular design with a graph generative model.
\newblock \emph{Chemical Science}, 11\penalty0 (4):\penalty0 1153--1164, 2020.

\bibitem[Maziarz et~al.(2021)Maziarz, Jackson-Flux, Cameron, Sirockin,
  Schneider, Stiefl, and Brockschmidt]{scaffold-based-gvae-arxiv21}
Krzysztof Maziarz, Henry Jackson-Flux, Pashmina Cameron, Finton Sirockin,
  Nadine Schneider, Nikolaus Stiefl, and Marc Brockschmidt.
\newblock Learning to extend molecular scaffolds with structural motifs.
\newblock \emph{arXiv preprint arXiv:2103.03864}, 2021.

\bibitem[Ar{\'u}s-Pous et~al.(2020)Ar{\'u}s-Pous, Patronov, Bjerrum, Tyrchan,
  Reymond, Chen, and Engkvist]{scaffold-based-lstm-attn-jcheminfo20}
Josep Ar{\'u}s-Pous, Atanas Patronov, Esben~Jannik Bjerrum, Christian Tyrchan,
  Jean-Louis Reymond, Hongming Chen, and Ola Engkvist.
\newblock Smiles-based deep generative scaffold decorator for de-novo drug
  design.
\newblock \emph{Journal of cheminformatics}, 12:\penalty0 1--18, 2020.

\bibitem[Langevin et~al.(2020)Langevin, Minoux, Levesque, and
  Bianciotto]{scaffold-constrained-rnn-jcim20}
Maxime Langevin, Herv{\'e} Minoux, Maximilien Levesque, and Marc Bianciotto.
\newblock Scaffold-constrained molecular generation.
\newblock \emph{Journal of Chemical Information and Modeling}, 2020.

\bibitem[He et~al.(2021{\natexlab{b}})He, Mattsson, Forsberg, Bjerrum,
  Engkvist, Tyrchan, Czechtizky, et~al.]{mmp-transformer-iclrw21}
Jiazhen He, Felix Mattsson, Marcus Forsberg, Esben~Jannik Bjerrum, Ola
  Engkvist, Christian Tyrchan, Werngard Czechtizky, et~al.
\newblock Transformer neural network for structure constrained molecular
  optimization.
\newblock 2021{\natexlab{b}}.

\bibitem[Wang et~al.(2019)Wang, Guo, Wang, Sun, and Huang]{smiles-bert-bcb19}
Sheng Wang, Yuzhi Guo, Yuhong Wang, Hongmao Sun, and Junzhou Huang.
\newblock Smiles-bert: large scale unsupervised pre-training for molecular
  property prediction.
\newblock In \emph{Proceedings of the 10th ACM international conference on
  bioinformatics, computational biology and health informatics}, pages
  429--436, 2019.

\bibitem[Rong et~al.(2020)Rong, Bian, Xu, Xie, Wei, Huang, and
  Huang]{grover-neurips20}
Yu~Rong, Yatao Bian, Tingyang Xu, Weiyang Xie, Ying Wei, Wenbing Huang, and
  Junzhou Huang.
\newblock Self-supervised graph transformer on large-scale molecular data.
\newblock \emph{Advances in Neural Information Processing Systems}, 33, 2020.

\bibitem[Schwaller et~al.(2019)Schwaller, Laino, Gaudin, Bolgar, Hunter, Bekas,
  and Lee]{reaction-pred-acs-cs19}
Philippe Schwaller, Teodoro Laino, Th{\'e}ophile Gaudin, Peter Bolgar,
  Christopher~A Hunter, Costas Bekas, and Alpha~A Lee.
\newblock Molecular transformer: a model for uncertainty-calibrated chemical
  reaction prediction.
\newblock \emph{ACS central science}, 5\penalty0 (9):\penalty0 1572--1583,
  2019.

\bibitem[Karpov et~al.(2019)Karpov, Godin, and
  Tetko]{retrosynthesis-transformer-icann19}
Pavel Karpov, Guillaume Godin, and Igor~V Tetko.
\newblock A transformer model for retrosynthesis.
\newblock In \emph{International Conference on Artificial Neural Networks},
  pages 817--830. Springer, 2019.

\bibitem[Elnaggar et~al.(2020)Elnaggar, Heinzinger, Dallago, Rihawi, Wang,
  Jones, Gibbs, Feher, Angerer, Bhowmik, and Rost]{prottrans-biorxiv20}
Ahmed Elnaggar, Michael Heinzinger, Christian Dallago, Ghalia Rihawi, Yu~Wang,
  Llion Jones, Tom Gibbs, Tamas Feher, Christoph Angerer, Debsindhu Bhowmik,
  and Burkhard Rost.
\newblock Prottrans: Towards cracking the language of life{\textquoteright}s
  code through self-supervised deep learning and high performance computing.
\newblock \emph{bioRxiv}, 2020.
\newblock \doi{10.1101/2020.07.12.199554}.
\newblock URL
  \url{https://www.biorxiv.org/content/early/2020/07/12/2020.07.12.199554}.

\bibitem[Grechishnikova(2021)]{protein-transformer-nature-sr21}
Daria Grechishnikova.
\newblock Transformer neural network for protein-specific de novo drug
  generation as a machine translation problem.
\newblock \emph{Scientific reports}, 11\penalty0 (1):\penalty0 1--13, 2021.

\bibitem[Rajan et~al.(2021)Rajan, Zielesny, and Steinbeck]{rajan2021stout}
Kohulan Rajan, Achim Zielesny, and Christoph Steinbeck.
\newblock Stout: Smiles to iupac names using neural machine translation.
\newblock \emph{Journal of Cheminformatics}, 13\penalty0 (1):\penalty0 1--14,
  2021.

\bibitem[Handsel et~al.(2021)Handsel, Matthews, Knight, and
  Coles]{handsel2021translating}
Jennifer Handsel, Brian Matthews, Nicola Knight, and Simon Coles.
\newblock Translating the molecules: adapting neural machine translation to
  predict iupac names from a chemical identifier.
\newblock 2021.

\bibitem[Krasnov et~al.(2021)Krasnov, Khokhlov, Fedorov, and
  Sosnin]{krasnov2021struct2iupac}
Lev Krasnov, Ivan Khokhlov, Maxim Fedorov, and Sergey Sosnin.
\newblock Struct2iupac--transformer-based artificial neural network for the
  conversion between chemical notations.
\newblock 2021.

\bibitem[Favre and Powell(2013)]{favre2013nomenclature}
Henri~A Favre and Warren~H Powell.
\newblock \emph{Nomenclature of organic chemistry: IUPAC recommendations and
  preferred names 2013}.
\newblock Royal Society of Chemistry, 2013.

\bibitem[Mikolov et~al.(2013)Mikolov, Chen, Corrado, and Dean]{word2vec}
Tomas Mikolov, Kai Chen, Greg Corrado, and Jeffrey Dean.
\newblock Efficient estimation of word representations in vector space.
\newblock \emph{arXiv preprint arXiv:1301.3781}, 2013.

\bibitem[Kim et~al.(2016)Kim, Thiessen, Bolton, Chen, Fu, Gindulyte, Han, He,
  He, Shoemaker, et~al.]{kim2016pubchem}
Sunghwan Kim, Paul~A Thiessen, Evan~E Bolton, Jie Chen, Gang Fu, Asta
  Gindulyte, Lianyi Han, Jane He, Siqian He, Benjamin~A Shoemaker, et~al.
\newblock Pubchem substance and compound databases.
\newblock \emph{Nucleic acids research}, 44\penalty0 (D1):\penalty0
  D1202--D1213, 2016.

\bibitem[Lowe et~al.(2011)Lowe, Corbett, Murray-Rust, and Glen]{opsin}
Daniel~M Lowe, Peter~T Corbett, Peter Murray-Rust, and Robert~C Glen.
\newblock Chemical name to structure: Opsin, an open source solution, 2011.

\bibitem[Donahue et~al.(2020)Donahue, Lee, and Liang]{donahue2020ilm}
Chris Donahue, Mina Lee, and Percy Liang.
\newblock Enabling language models to fill in the blanks.
\newblock In \emph{ACL}, 2020.

\bibitem[Ghose et~al.(1999)Ghose, Viswanadhan, and
  Wendoloski]{ghose1999knowledge}
Arup~K Ghose, Vellarkad~N Viswanadhan, and John~J Wendoloski.
\newblock A knowledge-based approach in designing combinatorial or medicinal
  chemistry libraries for drug discovery. 1. a qualitative and quantitative
  characterization of known drug databases.
\newblock \emph{Journal of combinatorial chemistry}, 1\penalty0 (1):\penalty0
  55--68, 1999.

\bibitem[Veber et~al.(2002)Veber, Johnson, Cheng, Smith, Ward, and
  Kopple]{veber2002molecular}
Daniel~F Veber, Stephen~R Johnson, Hung-Yuan Cheng, Brian~R Smith, Keith~W
  Ward, and Kenneth~D Kopple.
\newblock Molecular properties that influence the oral bioavailability of drug
  candidates.
\newblock \emph{Journal of medicinal chemistry}, 45\penalty0 (12):\penalty0
  2615--2623, 2002.

\bibitem[Hitchcock and Pennington(2006)]{hitchcock2006structure}
Stephen~A Hitchcock and Lewis~D Pennington.
\newblock Structure- brain exposure relationships.
\newblock \emph{Journal of medicinal chemistry}, 49\penalty0 (26):\penalty0
  7559--7583, 2006.

\bibitem[Bhal et~al.(2007)Bhal, Kassam, Peirson, and Pearl]{bhal2007rule}
Sanjivanjit~K Bhal, Karim Kassam, Ian~G Peirson, and Greg~M Pearl.
\newblock The rule of five revisited: applying log d in place of log p in
  drug-likeness filters.
\newblock \emph{Molecular pharmaceutics}, 4\penalty0 (4):\penalty0 556--560,
  2007.

\bibitem[Kudo and Richardson(2018)]{sentencepiece}
Taku Kudo and John Richardson.
\newblock Sentencepiece: A simple and language independent subword tokenizer
  and detokenizer for neural text processing.
\newblock In \emph{EMNLP (Demonstration)}, 2018.

\bibitem[Lacoste et~al.(2019)Lacoste, Luccioni, Schmidt, and
  Dandres]{emissionscalculator}
Alexandre Lacoste, Alexandra Luccioni, Victor Schmidt, and Thomas Dandres.
\newblock Quantifying the carbon emissions of machine learning.
\newblock \emph{arXiv preprint arXiv:1910.09700}, 2019.

\bibitem[Wolf et~al.(2019)Wolf, Debut, Sanh, Chaumond, Delangue, Moi, Cistac,
  Rault, Louf, Funtowicz, et~al.]{wolf2019huggingface}
Thomas Wolf, Lysandre Debut, Victor Sanh, Julien Chaumond, Clement Delangue,
  Anthony Moi, Pierric Cistac, Tim Rault, R{\'e}mi Louf, Morgan Funtowicz,
  et~al.
\newblock Huggingface's transformers: State-of-the-art natural language
  processing.
\newblock \emph{arXiv preprint arXiv:1910.03771}, 2019.

\bibitem[Paszke et~al.(2019)Paszke, Gross, Massa, Lerer, Bradbury, Chanan,
  Killeen, Lin, Gimelshein, Antiga, et~al.]{paszke2019pytorch}
Adam Paszke, Sam Gross, Francisco Massa, Adam Lerer, James Bradbury, Gregory
  Chanan, Trevor Killeen, Zeming Lin, Natalia Gimelshein, Luca Antiga, et~al.
\newblock Pytorch: An imperative style, high-performance deep learning library.
\newblock \emph{arXiv preprint arXiv:1912.01703}, 2019.

\bibitem[Pedregosa et~al.(2011)Pedregosa, Varoquaux, Gramfort, Michel, Thirion,
  Grisel, Blondel, Prettenhofer, Weiss, Dubourg, et~al.]{pedregosa2011scikit}
Fabian Pedregosa, Ga{\"e}l Varoquaux, Alexandre Gramfort, Vincent Michel,
  Bertrand Thirion, Olivier Grisel, Mathieu Blondel, Peter Prettenhofer, Ron
  Weiss, Vincent Dubourg, et~al.
\newblock Scikit-learn: Machine learning in python.
\newblock \emph{the Journal of machine Learning research}, 12:\penalty0
  2825--2830, 2011.

\end{thebibliography}

\newpage

\appendix

\section{Comparison To Prior Methods}
\label{app:related}
\begin{table}\noindent\begin{minipage}{\textwidth}
    \centering
    \begin{tabularx}{0.77\textwidth}{cccccc} \toprule
         \textbf{Method} &
         \textbf{Base Repr.} &
         \textbf{Model} &
         \textbf{T?}\footnote{Whether or not the model makes targeted modifications -- \emph{i.e.} whether the user can specify which part of the molecule the model should modify.} &
         \textbf{UD?}\footnote{Whether or not the model can train without using paired molecular data.} & 
         \textbf{CO?}\footnote{Whether or not the model can switch objectives without re-training or re-optimizing.} \\
         \midrule
         C5T5                                        & IUPAC        & T5 & \ch & \ch & \ch \\
         \cite{mmp-transformer-jcheminfo21}          & SMILES       & Seq2Seq/Transformer & \x & \x & \ch \\
         \cite{scaffold-constrained-rnn-jcim20}      & SMILES       & Any & \ch & \x & \x \\
         \cite{scaffold-based-lstm-attn-jcheminfo20} & SMILES       & LSTM & \ch & \x & \x \\
         \cite{vjtnn-iclr19}                         & Graph/Motifs & JT-VAE+GAN & \x & \x & \x \\
         \cite{deepscaffold-jcim19}                  & Graph/Atoms  & GNN+VAE & \x\footnote{The authors propose filtering results based on where the user wants to add a side chain, but the method itself does not target specific attachment points.} & \x & \x \\
         \cite{scaffold-based-vae-rsc-cs20}          & Graph/Atoms  & VAE+GNN & \x & \x & \ch \\
         \cite{scaffold-based-gvae-arxiv21}          & Graph/Motifs & VAE+GNN & \x & \ch & \ch \\
         \cite{liggpt-chemrxiv21}                    & SMILES       & GPT & \x & \ch & \ch \\
         \cite{cmg-transformer-chil21}               & SMILES       & Transformer+LSTM & \x & \x & \ch \\
         \cite{mmp-transformer-iclrw21}              & SMILES       & Transformer & \ch & \x & \ch \\
         \cite{crnn-nature-mi20}                     & SMILES & cRNN  & \x & \ch & \ch \\
         \cite{bombarelli-acs-cs18}                  & SMILES & VAE+ConvNet+GRU  & \x & \ch & \x \\
         \cite{condvae-jcheminfo18}                  & SMILES & cVAE  & \x & \ch & \ch \\
         \cite{vae-transformer-cs21}                 & SMILES & VAE+Transformer & \x & \ch & \x \\
         \cite{cgvae-neurips18}                      & Graph/Atoms  & VAE+GNN & \x & \ch & \x \\
         \cite{gcpn-neurips18}                       & Graph/Atoms & GCN+GAN+RL & \x & \ch & \x \\
         \cite{jtvae-icml18}                         & Graph/Motifs & JT-VAE & \x & \ch & \x \\
         \cite{cyclegan-jchi20}                      & Graph/Motifs & CycleGAN+JT-VAE & \x & \ch & \ch \\
         \cite{graphaf-iclr20}                       & Graph/Atoms & Flow+RL & \x & \ch & \x \\
         \cite{olivecrona2017molecular}              & SMILES & RNN+RL & \x & \ch & \x \\
         \bottomrule \\
    \end{tabularx}
    \caption{Comparison of C5T5 to prior methods}
    \label{tab:related}
\end{minipage}\end{table}

Table \ref{tab:related} shows how C5T5 and prior methods for molecular optimization differ along several axes.

\section{Additional Experiments}
\label{app:novelty}
Section \ref{subsec:results-modifies} shows that C5T5 successfully modifies property values.
Here, for molecules generated to optimize logP, we show the novelty and validity of the generated molecules, along with a comparison to a baseline of the best eligible compound in PubChem.
To match how the model was trained and how new molecules were generated, eligible compounds are those that could be generated by masking any consecutive span of at most 5 IUPAC name tokens and replacing the masked tokens with any number of replacement tokens.\footnote{For computational efficiency, we filter out molecules that differ in length by more than 15 tokens, or that have more than 15 non-overlapping tokens in their bag of tokens, before checking whether they could indeed be generated by masking some length-5 sequence of tokens.}
Results are shown in Tables \ref{tab:novelty} and \ref{tab:lookup}.
Percent validity is the fraction of generated molecules that T5 generated with valid sentinel tokens that were considered chemically valid by the ChemAxon logP calculator.
Percent novelty is the fraction of distinct generated molecules that do not appear in PubChem (excluding when C5T5 re-generated the source molecule).
As shown, despite the comparative difficulty of learning IUPAC name syntax compared to SMILES syntax, C5T5 consistently finds novel and valid molecules that significantly outperform the best-in-dataset baseline.

\begin{table}[h]
    \centering
    \caption{For each source molecule, we show the percent of generated molecules that are novel (not in PubChem, including invalid names) and valid (can be parsed by ChemAxon), the number of generated molecules, the min/max logP of any generated molecule, the number of eligible compounds in PubChem and the min/max logP of all molecules in PubChem that could be generated by masking up to 5 consecutive tokens. IUPAC names of source molecules are listed in Table \ref{tab:lookup}.}
    \label{tab:novelty}
    \begin{tabular}{ccccccccc}
    \toprule
    \textbf{Src.} & \textbf{\# gen.} & \textbf{\% novel} & \textbf{\% valid} & \textbf{max gen.} & \textbf{min gen.} & \textbf{\# elig.} & \textbf{max PC} & \textbf{min PC} \\
    \midrule
1 & 82 & 95.1\% & 81.7\% & 14.22 & -5.04 & 26 & 10.81 & -2.7 \\
2 & 133 & 93.2\% & 91.7\% & 9.41 & -3.82 & 28 & 1.92 & -1.46 \\
3 & 217 & 98.6\% & 91.7\% & 9.15 & -2.17 & 4 & 3.01 & 1.66 \\
4 & 140 & 100.0\% & 94.3\% & 10.21 & 1.15 & 3 & 8.18 & 3.78 \\
5 & 128 & 92.2\% & 88.3\% & 6.91 & 0.86 & 19 & 4.24 & 2.35 \\
6 & 160 & 100.0\% & 75.6\% & 8.16 & 0.47 & 4 & 2.56 & 1.6 \\
7 & 159 & 99.4\% & 86.2\% & 9.05 & -2.22 & 8 & 3.91 & -1.34 \\
8 & 137 & 99.3\% & 84.7\% & 14.74 & -2.97 & 1 & 1.08 & 1.08 \\
9 & 122 & 95.1\% & 81.1\% & 8.57 & -4.89 & 34 & 4.75 & -0.07 \\
10 & 112 & 92.9\% & 88.4\% & 8.44 & -2.96 & 112 & 5.03 & -2.3 \\
11 & 127 & 97.6\% & 81.1\% & 9.52 & 0.11 & 9 & 5.13 & 2.28 \\
12 & 114 & 90.4\% & 85.1\% & 7.71 & -3.32 & 515 & 6.13 & -2.95 \\
13 & 115 & 94.8\% & 90.4\% & 8.12 & -1.91 & 36 & 4.05 & 0.3 \\
14 & 149 & 97.3\% & 85.2\% & 14.09 & -4.3 & 7 & 2.1 & 0.44 \\
15 & 135 & 100.0\% & 83.0\% & 7.52 & -0.7 & 2 & 5.08 & 4.12 \\
16 & 214 & 100.0\% & 97.7\% & 6.86 & -1.83 & 3 & 1.91 & -0.05 \\
17 & 156 & 99.4\% & 92.3\% & 10.58 & -2.88 & 2 & 0.76 & 0.76 \\
18 & 231 & 98.7\% & 83.1\% & 9.79 & -1.35 & 6 & 5.6 & 3.46 \\
19 & 148 & 93.2\% & 82.4\% & 9.98 & -0.19 & 63 & 7.01 & 0.76 \\
20 & 143 & 96.5\% & 72.7\% & 14.12 & 0.33 & 14 & 4.61 & 1.31 \\
21 & 232 & 99.1\% & 85.3\% & 9.9 & -1.6 & 6 & 3.15 & 1.14 \\
22 & 150 & 96.0\% & 92.7\% & 7.96 & -0.74 & 22 & 4.95 & 2.71 \\
23 & 127 & 94.5\% & 87.4\% & 7.54 & -3.1 & 18 & 4.67 & 0.37 \\
24 & 160 & 99.4\% & 82.5\% & 7.54 & -1.53 & 2 & 1.28 & 1.28 \\
25 & 274 & 100.0\% & 95.6\% & 10.82 & 1.12 & 2 & 6.12 & 5.82 \\
    \bottomrule
    \end{tabular}
\end{table}

\begin{table}[h]
    \small
    \centering
    \caption{IUPAC Name lookup table for Table \ref{tab:novelty}}
    \label{tab:lookup}
    \begin{tabular}{cl}
    \toprule
    \textbf{ID} & \textbf{Source Molecule} \\
    \midrule
1 & 3,3-bis(aminomethyl)pentane-1,5-diol \\
2 & 1-(3-hydroxypropyl)-N-(1-methoxybutan-2-yl)pyrazole-4-sulfonamide \\
3 & 4-chloro-N-[2-[[2-(4-fluorophenyl)acetyl]amino]ethyl]-1,3-thiazole-5-carboxamide \\
4 & 1-[(1S)-1-(3-fluorophenyl)propyl]-3-iodoindole \\
5 & 4-(4-fluorophenyl)-N-[(1R,2R)-2-methylcyclohexyl]piperazine-1-carbothioamide \\
6 & N'-(3-ethyl-4-oxophthalazine-1-carbonyl)-4-methyl-2-phenyl-1,3-thiazole-5-carbohydrazide \\
7 & (E)-2-methoxy-3-methylhex-4-en-1-ol \\
8 & N-methyl-1-[2-(4-methylthiadiazol-5-yl)-1,3-thiazol-4-yl]methanamine \\
9 & 2-[(7-methyl-[1,2,4]triazolo[1,5-a]pyridin-2-yl)amino]ethylurea \\
10 & 4-[[2-(2-oxopyridin-1-yl)acetyl]amino]benzoic acid \\
11 & [6-prop-2-enoxy-4-(trifluoromethyl)pyridin-2-yl]hydrazine \\
12 & 4-(2-methylphenyl)sulfonylpiperidin-3-amine \\
13 & 3-[ethyl(2-methylpropyl)amino]propane-1-thiol \\
14 & 6-methoxy-4-N-methyl-4-N-[(2-methylfuran-3-yl)methyl]pyrimidine-4,5-diamine \\
15 & 3-phenylmethoxy-5-(trifluoromethoxy)quinoline-2-carboxylic acid \\
16 & 3-[4-[acetamido-[3-methoxy-4-[(2-methylphenyl)carbamoylamino]phenyl]methyl]piperidin-1-yl]-\\&3-phenylpropanoic acid \\
17 & (3R)-3-[[(2S)-2-[benzyl(methyl)amino]butanoyl]amino]pyrrolidine-1-carboxamide \\
18 & 6-cyclobutyl-2-N-[3-(1-ethylsulfinylethyl)phenyl]-5-(trifluoromethyl)pyrimidine-2,4-diamine \\
19 & 6-fluoro-2-(4-phenylpyridin-2-yl)-1H-benzimidazole \\
20 & 4-chloro-3-(2-oxo-1,3-dihydroindol-5-yl)benzonitrile \\
21 & 1-(6-tert-butylpyridazin-3-yl)-N-methyl-N-[(2-methyl-1,3-oxazol-4-yl)methyl]azetidin-3-amine \\
22 & 2-[(4aR,8aS)-3,4,4a,5,6,7,8,8a-octahydro-1H-isoquinolin-2-yl]-N-(2,4-dimethoxyphenyl)acetamide \\
23 & 2-[2-(2,4-dichlorophenoxy)ethoxy]-4-methoxybenzoic acid \\
24 & (2Z)-2-[(1,7-dimethylquinolin-1-ium-2-yl)methylidene]-1-ethyl-7-methylquinoline \\
25 & N'-[(3S)-1-[[3-(2,4-dichlorophenyl)phenyl]methyl]-2-oxoazepan-3-yl]-\\&3-(2-methylpropyl)-2-prop-2-enylbutanediamide \\
\bottomrule
    \end{tabular}
    \label{table:lookup}
\end{table}

\section{Experimental Details}
\label{app:experiment_details}
\paragraph{Dataset Preparation}
We download PubChem from \url{ftp.ncbi.nlm.nih.gov/pubchem/Compound/CURRENT-Full/XML/} and extracted each molecule's Preferred IUPAC Name and computed octanol-water partition coefficient (logP).
There are 109M total compounds in the version of PubChem we downloaded in January 2021.
For experiments using logP, we used the XLogP3 values from PubChem, which were computed using OpenEye's software. 
For logD (pH=7), refractivity, and polar surface area, we computed values using ChemAxon's calculator with default parameters.
Separately for each target property, we excluded chemicals that had no logP value in PubChem or that were not parseable by ChemAxon's calculator.
Of the remaining molecules, we randomly split into a training set with 90M compounds and a validation set with $\sim$10M-19M compounds.

\paragraph{Tokenization}
We use HuggingFace's T5Tokenizer, which is based on the SentencePiece algorithm \cite{sentencepiece}.
Because our goal is to have tokens that domain experts are familiar with, we do not train SentencePiece on the IUPAC names, since doing so learns both combinations of and substrings of moiety names.
Instead, we manually specify the tokens to be all the keywords in the Opsin IUPAC name parsing library \cite{opsin}.
To these keywords, we add locants 1--100, stereochemistry markers (R, S, E, Z,...), and a few miscellaneous tokens to e.g. handle spiro centers.
This leads to a total of 1274 tokens.
After tokenization, we truncate all names to at most 128 tokens.

\paragraph{Training}
We train a t5-large model ($\sim$700M params) available from HuggingFace that was pretrained on English text.
We keep the first 1274 embeddings from the pretrained embedding table, along with the pretrained embeddings for the 100 sentinel tokens.
When training, we mask 15\% of the tokens in each input in spans of mean length 3 tokens, with a minimum span length of 1.
We use a linear warmup of the learning rate for 10,000 steps followed by a $1/T$ decay.
All models were trained using the AdamW optimizer.
We train the logP model for 2.5M iterations using a max learning rate of $10^{-3}$.
We train the refractivity/logD/Polar SA models using a maximum learning rate of $2\times 10^{-4}$/$10^{-4}$/$2\times 10^{-4}$ starting from the logP model after 1M/2.5M/2.5M iterations.
(We trained using the latest model available when we started a run.)
All models were trained on 8 NVIDIA A100s with a batch size of 16 per GPU.

\paragraph{Generation}
We generate novel molecules greedily from the output of T5's decoder.
We discard any generations where the sentinel tokens do not line up, and we further discard any molecules that cannot be parsed by ChemAxon's calculators.

\paragraph{Cloud Computing Cost}
We train T5-Large models \cite{t5} on PubChem for each investigated property value using AWS p4d.24xlarge instances in the us-east-1 region. 
The logP model was trained for 2.5M iterations over 8 days; the logD and refractivity models for 350k iterations over 1 day each (initialized from the logP model after 2.5M iterations); and the Polar SA model for 1.6M iterations over 5 days (initialized from the logP model after 1M iterations).
Total on-demand AWS cost for the models presented here is $\sim$ \$12,000, and total carbon dioxide-equivalent emissions is $\sim$ 320 kg \cite{emissionscalculator}.\footnote{The emissions calculator does not yet support NVIDIA A100 GPUs, so we calculated with V100-32GB instead.}

\paragraph{Software}
We use  HuggingFace 4.2.2 (Apache 2.0 license) \citep{wolf2019huggingface}, PyTorch 1.8.0 (BSD) \citep{paszke2019pytorch}, ChemAxon 20.17.0 (Academic License), scikit-learn 0.24.2 (New BSD) \citep{pedregosa2011scikit} and python 3.9 (PSFL).

\end{document}